\documentclass[journal,twoside]{IEEEtran}

\usepackage{graphicx}
\usepackage{amsmath}
\usepackage{amssymb}
\usepackage[colorlinks]{hyperref}
\usepackage{verbatim}

\def\0{{\bf 0}}
\def\1{{\bf 1}}

\def\eg{{\em e.g.}}
\def\ie{{\em i.e.}}

\usepackage{tabularx} 
\usepackage{booktabs}   
\usepackage{mathrsfs} 
\usepackage{multirow} 
\usepackage{amsfonts} 
\usepackage{wrapfig}  
\usepackage{cite}
\usepackage{epstopdf}
\usepackage{textcomp}

\begin{document}

\title{DomainATM: Domain Adaptation Toolbox for Medical Data Analysis}

\author{Hao~Guan, and Mingxia~Liu,~\IEEEmembership{Senior Member,~IEEE}
\thanks{H. Guan and M. Liu are with the Department of Radiology and Biomedical Research Imaging Center, University of North Carolina at Chapel Hill, Chapel Hill, NC 27599, USA. Corresponding author: M. Liu (mxliu@med.unc.edu).}
\thanks{This work was supported in part by NIH grants (No.~AG073297 and AG041721).}
}


\maketitle

\begin{abstract}
Domain adaptation (DA) is an important technique for modern machine learning-based medical data analysis,  which aims at reducing distribution differences between different medical datasets. 
A proper domain adaptation method can significantly enhance the statistical power by pooling data acquired from multiple sites/centers.
To this end, we have developed the Domain Adaptation Toolbox for Medical data analysis (DomainATM) -- an open-source software package designed for fast facilitation and easy customization of domain adaptation methods for medical data analysis.
The DomainATM is implemented in MATLAB with a user-friendly graphical interface, and it consists of a collection of popular data adaptation algorithms that have been extensively applied to medical image analysis and computer vision.
With DomainATM, researchers are able to facilitate fast feature-level and image-level adaptation, visualization and performance evaluation of different adaptation methods for medical data analysis. 
More importantly, the DomainATM enables the users to develop and test their own adaptation methods through scripting,  greatly enhancing its utility and extensibility. 
An overview characteristic and usage of DomainATM is presented and illustrated with three example experiments, demonstrating its effectiveness, simplicity, and flexibility.
The software, source code, and manual are available online. 
\end{abstract}


%


\section{Introduction}

\IEEEPARstart{M}{edical} data analysis is nowadays being boosted by modern statistical analysis tools, \ie, machine learning~\cite{deo2015machine,rajkomar2019machine,barragan2021artificial,erickson2017machine,fatima2017survey}. 
Classic machine learning typically assumes that training dataset (source domain) and test dataset (target domain) follow an independent but identical distribution~\cite{learning}. 
In real-world practice, however, this assumption can hardly hold due to the 
well-known ``domain shift" problem~\cite{shift_1,pooch2020can,kondrateva2021domain}. 
In medical imaging, domain shift or data heterogeneity is widespread and caused by different scanning parameters (\ie, between-scanner variability) and subject populations in multiple imaging sites.
It may increase the test error along with the distribution difference between training and test data~\cite{test_error_1,test_error_2}. 
Thus the domain shift/difference may greatly degrade statistical power of multi-site/multi-center studies and hinder the building of effective machine learning models.

\begin{figure}[!tbp]
\setlength{\belowcaptionskip}{-2pt}
\setlength{\abovecaptionskip}{-2pt}
\setlength{\abovedisplayskip}{-2pt}
\setlength{\belowdisplayskip}{-2pt}
\center
 \includegraphics[width= 1.0\linewidth]{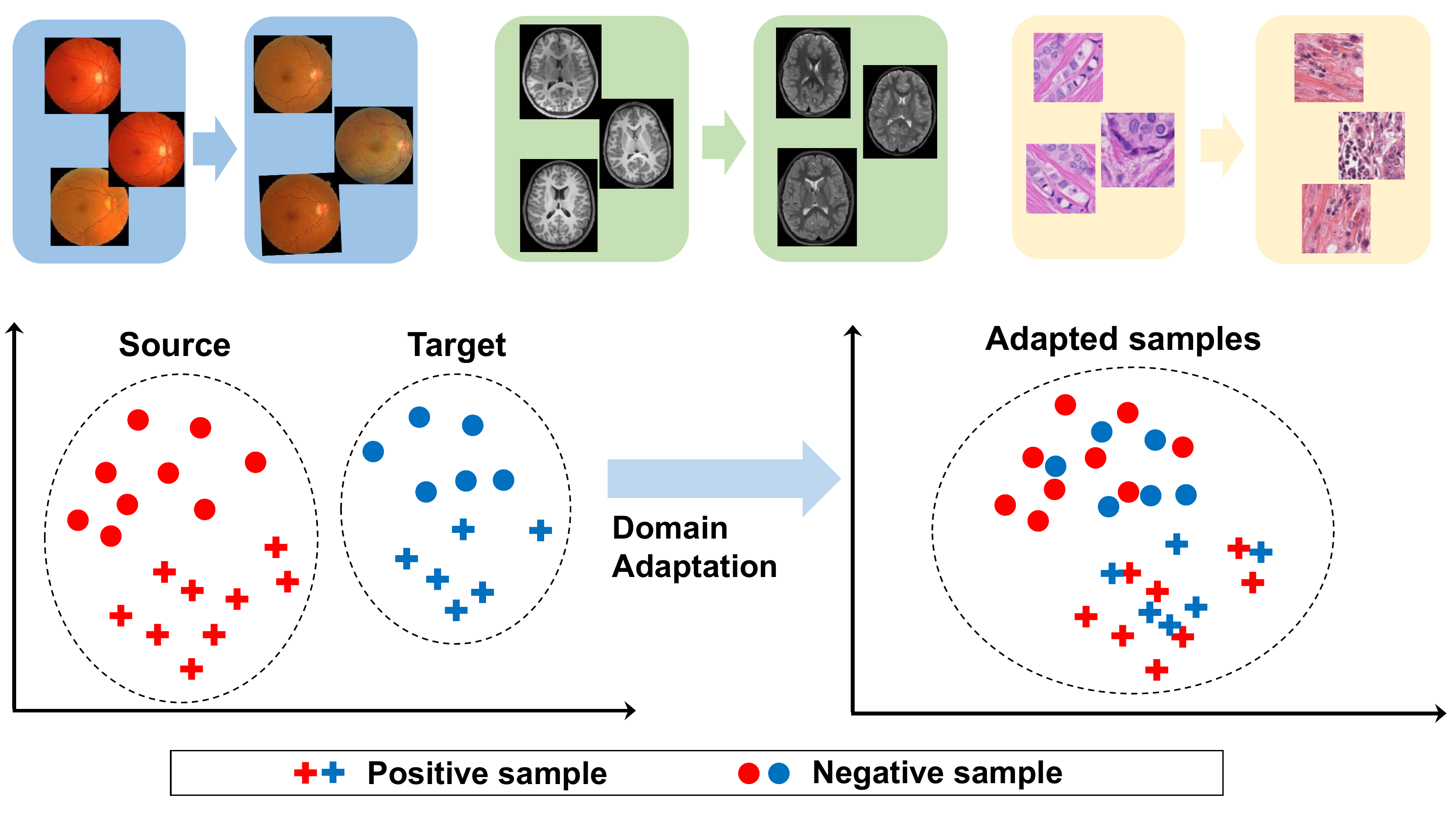}
 \caption{Illustration of the ``domain shift'' phenomenon~\cite{shift_1} (top row) and the fundamental of domain adaptation (distribution of source and target samples before and after adaptation).}
 \label{fig_DA}
\end{figure}
\begin{figure*}[t]
\setlength{\belowcaptionskip}{-2pt}
\setlength{\abovecaptionskip}{-2pt}
\setlength{\abovedisplayskip}{-2pt}
\setlength{\belowdisplayskip}{-2pt}
\center
 \includegraphics[width= 0.98\linewidth]{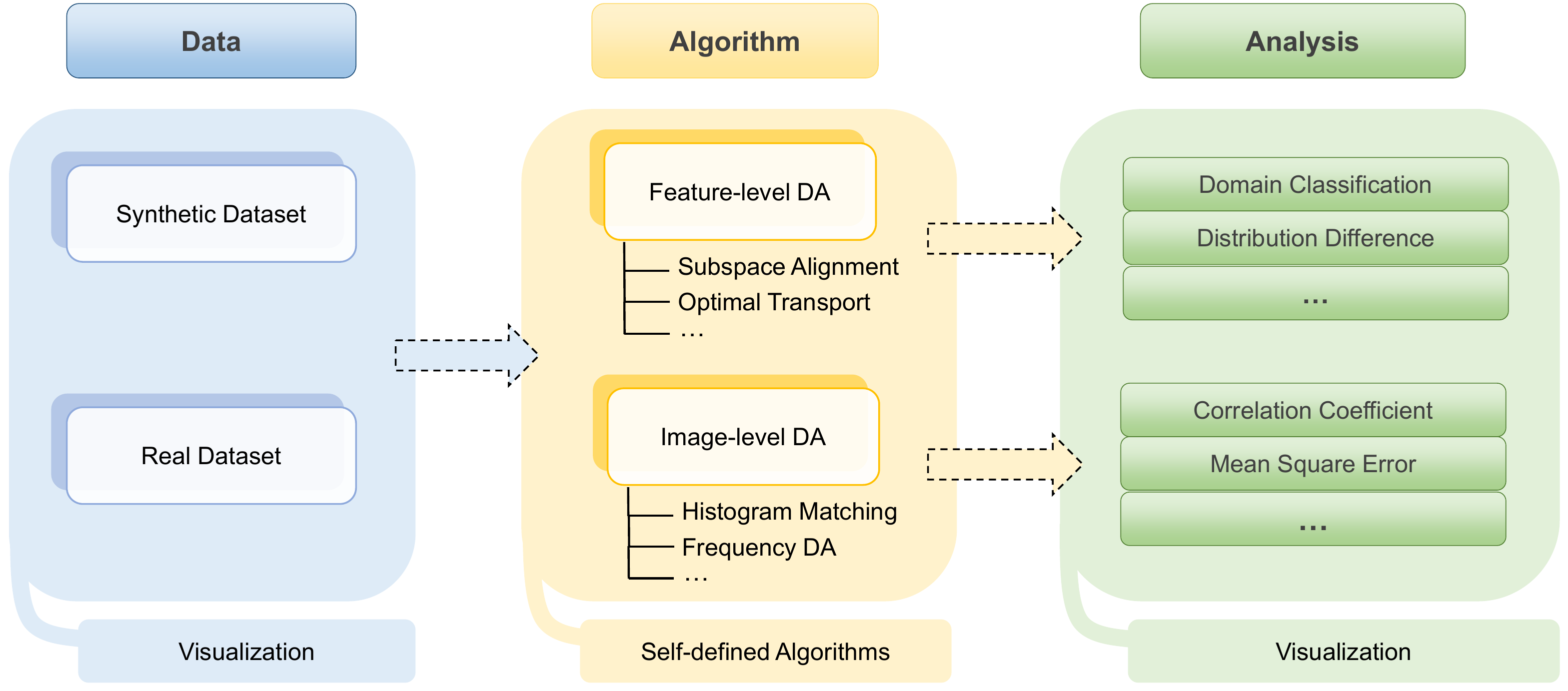}
 \caption{Illustration of workflow of the DomainATM software. The DomainATM consists of three major components: 1) the data module loads or creates the datasets; 2) the algorithm module conducts feature-level or image-level domain adaptation and saves the results; and 3) the evaluation module assesses the adaptation performance according to specific metrics. DA: Domain Adaptation.}
 \label{fig_workflow}
\end{figure*}

For handling the domain shift problem among datasets and enhancing the generalization ability of machine learning models, domain adaptation has gradually 
come under the spotlight of the research community~\cite{DA_1,DA_2,DA_3,kouw2019review,patel2015visual}.
In the field of medical data analysis, domain adaptation has gained considerable attention and increasing interest recently~\cite{DA_medical,valverde2021transfer}.
Briefly, domain adaptation can be defined as follows.
Let $\mathcal{X} \times \mathcal{Y}$ represent the joint feature space of samples and their corresponding category labels. 
A source domain  $\mathcal{S}$ and a target domain  $\mathcal{T}$ are defined on the joint feature space, with different distributions $\mathbf{P_S}$ and $\mathbf{P_T}$, respectively. Suppose there are $n_s$ samples (subjects) with or without category labels in the source domain, as well as $n_t$ samples in the target domain without category labels.  
Then the problem is how to reduce the distribution differences/variability between source and target domains so as to increase the performance of down-streaming tasks such as classification or segmentation.

Many domain adaptation methods have been proposed or utilized in the field of medical data analysis which shows  tremendous applicability.
Most solutions, however, are implemented independently for very specific scenarios or target applications. Researchers often need to re-implement an algorithm or do methodological tailoring. 
The differences in implementation will often cause inconsistent experiment and analysis results.
There is a lack of a unified platform for extensive comparison of different domain adaptation methods, helping avoiding hand-crafted re-implementation for specific medical data analysis research.
Thus a software toolbox that provides a platform of different adaptation methods is quite beneficial and necessary for researchers to compare, evaluate and select the proper method for their research project.

An important issue for medical imaging researchers is the fast facilitation of domain adaptation algorithms. Due to privacy protection issues, many real-world medical data sets are not accessible or with restrictions. 
Using synthetic data which is able to simulate the ``domain shift'' phenomenon in a machine learning setting will greatly boost the efficiency.
Another limitation is the complexity of certain domain adaptation methods. 
Time-consuming model training and exhaustive parameter tuning will be rather inconvenient, especially for researchers without high-level programming skills.
Thus, fast facilitation of domain adaptation methods with real-time visualization for performance check is beneficial for medical data analysis.

We also observe that in medical imaging image-level domain adaptation is an important topic~\cite{DA_medical}. For example, MRIs acquired from different scanners may negatively influence the analysis result~\cite{lee2019estimating,wittens2021inter}. This has become the concern of many radiologists and neuroscientists. Thus incorporating both feature-level and image-level adaptation methods into one platform is beneficial for related medical imaging research.  

In light of these motivations, we develop the Domain Adaptation Toolbox for Medical data analysis (DomainATM) -- a software package that offers a platform for simulating, evaluating and developing different domain adaptation algorithms for medical data analysis.
The toolbox is designed with a major principle that it could help researchers do fast facilitation of adaptation methods. 
Besides real-world medical data, synthetic data with user-defined statistical properties can be generated quickly for real-time simulation.
As for adaptation processing, both feature-level and image-level domain adaptation algorithms are included in the software package with a graphical-user-interface (GUI). 
The running results will be automatically saved which can be further analyzed by the evaluation module of the toolbox.
All the algorithms have consistent input/output formats under which the users can define their own data adaptation algorithms and add them to the DomainATM freely. Thus the toolbox has good flexibility and scalability.

This paper is organized as follows. 
In Section~\ref{Overview}, we introduce the characteristics of DomainATM, including its overall structure, key features and functions.
In Section~\ref{Workflow}, the workflow of DomainATM for facilitation of domain adaptation is described.
In Section~\ref{Algorithms}, representative domain adaptation methods that have been included in the toolbox are presented.
In Sections~\ref{Experiment1} and \ref{Experiment2}, experiments for both feature-level and image-level adaptation are conducted to illustrate the application of the toolbox.
This paper is concluded in Section~\ref{Conclusion}.

\section{Toolbox Overview/Characteristics} \label{Overview}
The main structure of the DomainATM is illustrated in Fig.~\ref{fig_workflow}.
Currently, the toolbox consists of three modules. 
1) The \textbf{data module} is responsible for loading and generating datasets. It can directly load an existing medical dataset (in {\em .mat} data file) or create synthetic datasets with user-defined statistical properties that can simulate domain shift. 
A dataset is in the format of $M \times N$ matrix, where ${M}$ denotes the number of samples while $N$ represents the feature dimension. 
2) The \textbf{algorithm module} contains the implementations of different domain adaptation methods. All these adaptation algorithms have uniform input/output parameter formats. Users can easily add their self-defined algorithms into the toolbox with the same input/output format. By default, several representative methods which have been widely used in medical data analysis are included in the DomainATM. 
These methods can be categorized into {\em feature-level}  adaptation methods and {\em image-level} adaptation methods.
Besides, inspired by the design philosophy of fast facilitation, most of the algorithms included in the toolbox can run in real time and output results in seconds.
3) The \textbf{evaluation module} assesses the performance of different adaptation methods. 
For feature-level adaptation methods, we employ two evaluation metrics, including: domain-level classification accuracy and domain distribution distance.  
For image-level adaptation methods, we use three evaluation metrics, including: correlation coefficient (CC), peak signal-noise ratio (PSNR) and mean square error (MSE).
The DomainATM provides visualization functions to visualize the data distribution (or images) before and after adaptation which helps investigate and understand the performance of different domain adaptation algorithms. 

\begin{figure}[!tbp]
\setlength{\belowcaptionskip}{-2pt}
\setlength{\abovecaptionskip}{-2pt}
\setlength{\abovedisplayskip}{-2pt}
\setlength{\belowdisplayskip}{-2pt}
\center
 \includegraphics[width= 0.90\linewidth]{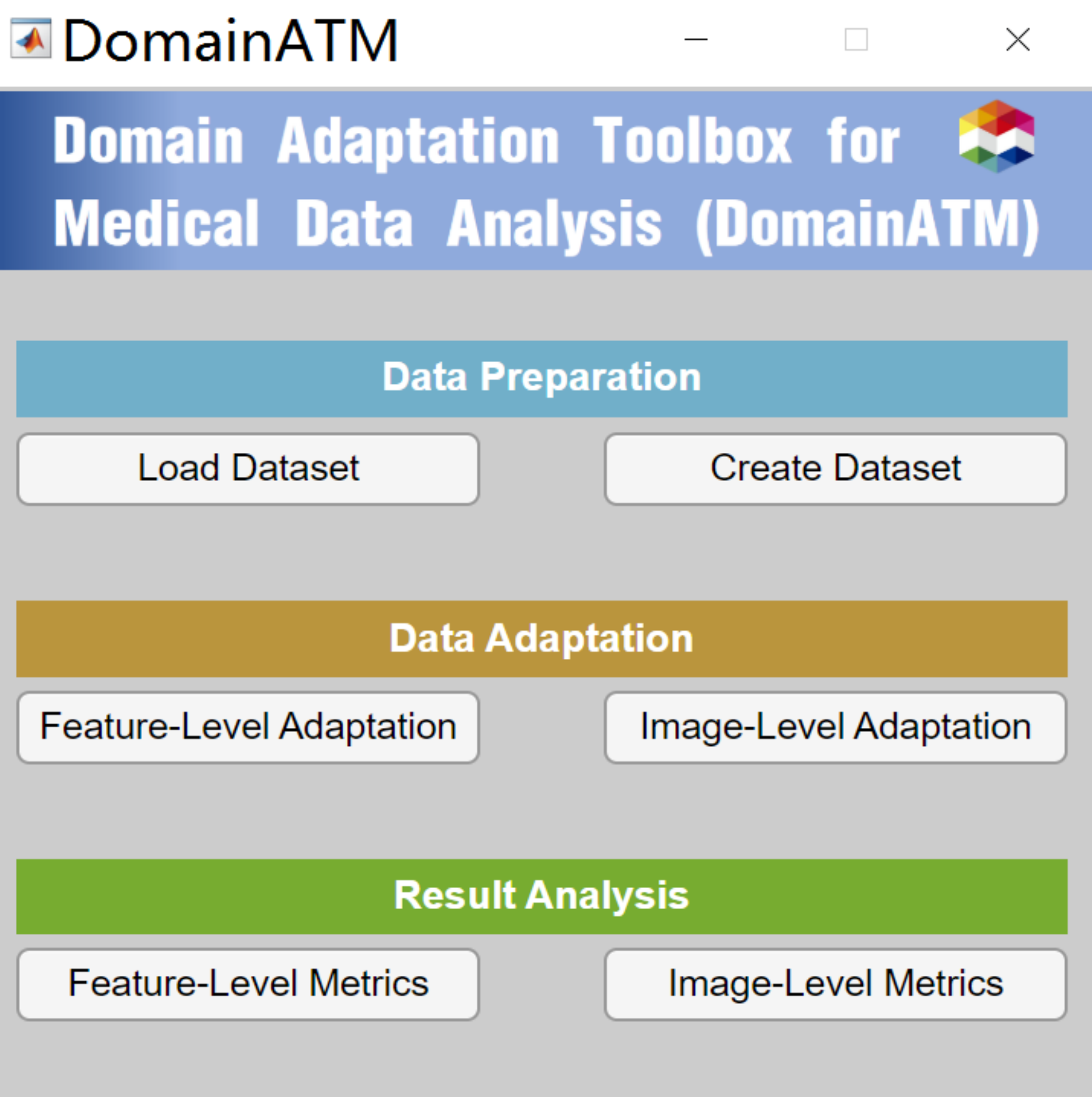}
 \caption{Graphical-User-Interface (GUI) of DomainATM. 
 }
 \label{fig_GUI}
\end{figure}

The DomainATM is implemented in MATLAB. 
It can be easily used with a graphical-user-interface (GUI), as shown in Fig.~\ref{fig_GUI}. Hardware platform can be a CPU-based PC without too many computation or memory resources. 
For advanced users, DomainATM provides an interface for writing MATLAB scripts to implement self-defined domain adaptation methods. 
The software, manual and source code for DomainATM are accessible online\footnote{https://mingxia.web.unc.edu/domainatm/}. 

\section{Toolbox Workflow} \label{Workflow}

\subsection{Creating/Loading Data}
The DomainATM can work for both feature-level adaptation and image-level adaptation.
These two key modules in the toolbox are independent to each other.
With respect to the input of feature-level adaptation, the toolbox accepts data in standard MATLAB {\em .mat} file format. Each row represents an observation (subject or sample) while every column represents a feature.
Existing real-world medical datasets (in \emph{.mat} format) can be directly imported and loaded into the toolbox for processing.
In addition, the users can create a synthetic dataset. After assigning the sample number, mean value and covariance matrix, the toolbox can automatically generate a synthetic dataset following normal distribution.
After loading the real/synthetic data, their distribution will be automatically displayed in the toolbox.
Both the real-world and created datasets are stored in the ``data'' subfolder of the toolbox.

For image-level adaptation, the toolbox currently accepts 3D volumetric data (in {\em .nii} format). 
All the data will be converted to inner-built data in MATLAB.
After loading the volumetric data, a middle slice (in axial view) will be automatically shown.

\subsection{Selecting Domain Adaptation Algorithms}
After loading the data, the following procedure is to select, configure, and run the domain adaptation methods.
Most adaptation methods have several hyper-parameters to be set. Users can tune them according to the specific tasks. Otherwise, default settings of these methods will be used.
After configuration, the users can run the algorithms. 
All the built-in methods provided by the toolbox are simple, easy to use, and can run in real time within 5 seconds (on a PC with an Intel i-7 CPU, 16 GB memory). 

After running the adaptation methods, the results will be automatically saved in the ``evaluation'' subfolder of the toolbox. 
For feature-level adaptation, the original source/target data, and the adapted source/target data will be saved (in \emph{.mat} data format).
For image-level adaptation, the adapted source image (target image is used as the reference image and will not be changed) will be saved (in \emph{.nii} format).
All the files are named with the corresponding adaptation method with time information as the suffix. 

\subsection{Evaluating Data Adaptation Performance}
After running the adaptation methods and getting the results, performance evaluation can be conducted for the methods.
For feature-level adaptation, we use {\em distribution difference} and {\em domain-level classification accuracy} as two metrics to assess the adaptation performance.
For image-level adaptation, we adopt {\em correlation coefficient (CC)}, {\em peak signal-to-noise ratio (PSNR)} and {\em mean-square error (MSE)} to evaluate the adaptation result.
More details about these evaluation metrics will be elaborated in the experiment section.

\subsection{Visualization of Data Adaptation Results}
Besides quantitative evaluation, result visualization is useful for qualitative analysis. 
The DomainATM provides visualization functions which help users better understand domain adaptation for medical data/images.
For feature-level adaption, the feature distribution (in 2D space) of data before and after adaptation can be visualized. 
High-dimensional features will be mapped to 2D feature space via t-SNE~\cite{tsne}.
For image-level adaptation, the adapted source image, the original source and target images can be viewed using the toolbox. After the adapted images have been saved in the ``evaluation'' subfolder, they can also be visually inspected by other medical imaging software.

\subsection{Extension: Adding Self-Defined Data Adaptation Algorithm}
In some tasks of medical data analysis, users might need to develop their own domain adaptation methods. 
The DomainATM supports self-defined algorithms for task-specific usage. 
The users can write a MATLAB script to define and implement their algorithms.
The input/output format of the self-defined functions has to be consistent with other built-in adaptation methods.
When adding an new algorithm, the self-defined script should be put in the ``algorithms\_feat" (feature-level) or the ``algorithms\_img" (image-level) subfolders in the toolbox.
One can simply run and analyze their methods like the other built-in ones through GUI. 

\section{Algorithms}  \label{Algorithms}
In this section, we briefly introduce the algorithms for feature-level and image-level data adaptation included in the DomainATM. 
More details on these algorithms can be found in the online manual.

\subsection{Feature-Level Data Adaptation Algorithm}
\subsubsection{Baseline}
No feature-level domain adaptation is utilized.
Both source and target data are kept in their original distributions (in the feature space).

\subsubsection{Subspace Alignment (SA)}
In this algorithm~\cite{SA}, the source and target medical data are represented by subspaces in terms of eigenvectors. The source data are projected to the target domain through a transformation matrix. No category labels of source domain are needed. The key hyper-parameter is the dimension of the shared subspace.

\subsubsection{Correlation Alignment (CORAL)}
In this algorithm~\cite{CORAL}, domain shift/difference is minimized by aligning the second-order statistics (\eg, covariance) of source and target distributions. No category label information and hyper-parameters are required for this method. 

\subsubsection{Transfer Component Analysis (TCA)}
In this algorithm~\cite{TCA}, a subspace shared by the source and target domain is searched in a reproducing kernel Hilbert space by minimizing the maximum mean discrepancy (MMD) distance. No source category labels are demanded. The key hyper-parameters are the kernel type and subspace dimension.

\subsubsection{Optimal Transport (OT)}
In this algorithm~\cite{OT}, the samples in the source domain are projected into the target domain while keeping their conditional distributions. The projection is facilitated through minimization of Wasserstein distance between the two distributions. No category labels of the source domain are used. The key hyper-parameter is the regularization coefficient.

\subsubsection{Joint Distribution Adaptation (JDA)}
In this algorithm~\cite{JDA}, maximum mean discrepancy (MMD) is adopted to measure domain distribution differences, and is integrated into Principal Component Analysis (PCA) to build feature representation that is robust to domain shift. Source category labels are needed in this algorithm. The key hyper-parameters include kernel type, subspace dimension and regularization parameter.

\subsubsection{Transfer Joint Matching (TJM)}
In this algorithm~\cite{TJM}, feature matching and instance reweighting strategies are combined to reduce domain shift. Minimization of maximum mean discrepancy (MMD) and $l_{2,1}$ norm sparsity penalty on source data are integrated into PCA to construct domain-invariant features.
Category labels of source domain are required. The key hyper-parameters include kernel type, subspace dimension and regularization parameter.

\subsubsection{Geodesic Flow Kernel (GFK)}
In this algorithm~\cite{GFK}, the source and target data are embedded into the Grassmann manifolds, and the geodesic flows between them are used to model domain shift. Domain adaptation is conducted by projecting the data into several domain-invariant subspaces on the geodesic flow. Source category labels can be either used or not. The key hyper-parameter is the subspace dimension.

\subsubsection{Scatter Component Analysis (SCA)}
In this algorithm~\cite{SCA}, original features are firstly projected to a reproducing kernel Hilbert space.  Domain adaptation is then conducted through an optimization formulation, including maximizing the class separability, maximizing the data separability, and minimizing domain mismatch.
Category labels of the source domain are used during adaptation. The key parameter is the dimension of the transformed space.

\subsubsection{Information-Theoretical Learning (ITL)}
In this algorithm~\cite{ITL}, an optimal feature space is learned through jointly maximizing domain similarity and minimizing the expected classification error
on target samples. Source category labels are required. The key hyper-parameters include subspace dimension and regularization parameter.

\subsection{Image-Level Data Adaptation Algorithm}
\subsubsection{Baseline}
For two medical images acquired by different scanners/sites, no domain adaptation is facilitated in this method. 
Instead, the homogeneity/heterogeneity of the paired original images is directly compared in terms of certain evaluation metrics.

\subsubsection{Histogram Matching (HM)}
This method transforms the source image to make its histogram matches the histogram derived from the target image~\cite{HM}.
After adaptation, the intensity distributions of the source and target images become closer.

\subsubsection{Spectrum Swapping-based Image-level Harmonization (SSIMH)}
In this method~\cite{SSIMH}, the source and target images are firstly transformed into the frequency domain (\eg, through Discrete Cosine Transform). Then, part of the low-frequency region of source image is replaced by the corresponding low-frequency area of the target image. Finally, the source image in the revised frequency domain is inverted back to the spatial domain to get the adapted image. 
The key hyper-parameter of this method is the threshold which defines the low-frequency region that is swapped between source and target images. In the toolbox, the default value is set to 3.
\begin{figure*}[t]
\center
 \includegraphics[width= 1.0\linewidth]{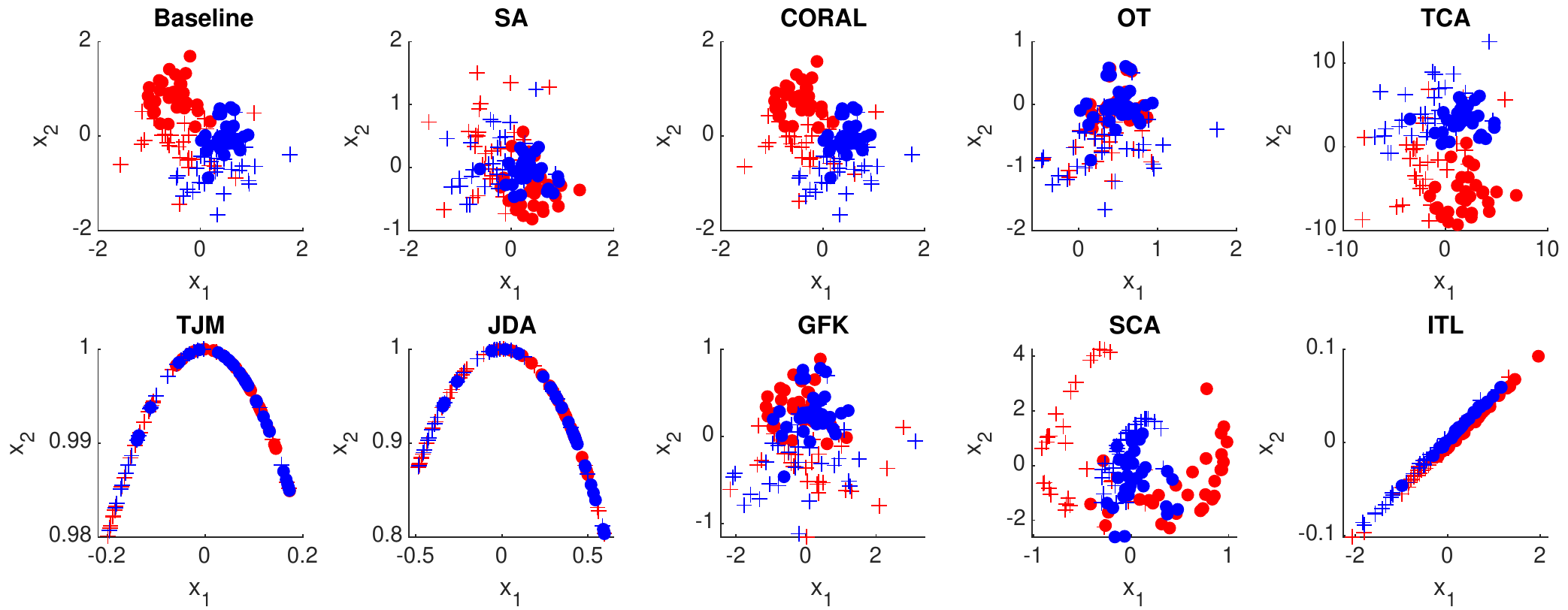}
 \caption{Distribution of the synthetic data (baseline) and adapted data by nine different domain adaptation methods in the DomainATM toolbox. ({\color{red}\textbf{+}} positive source samples; {\color{blue}\textbf{+}} positive target samples; {\color{red}\textbf{$\bullet$}} negative source samples; {\color{blue}\textbf{$\bullet$}} negative target samples)
 } 
 \label{fig_synthetic}
\end{figure*}

\begin{figure}[t]
\center
 \includegraphics[width= 1.0\linewidth]{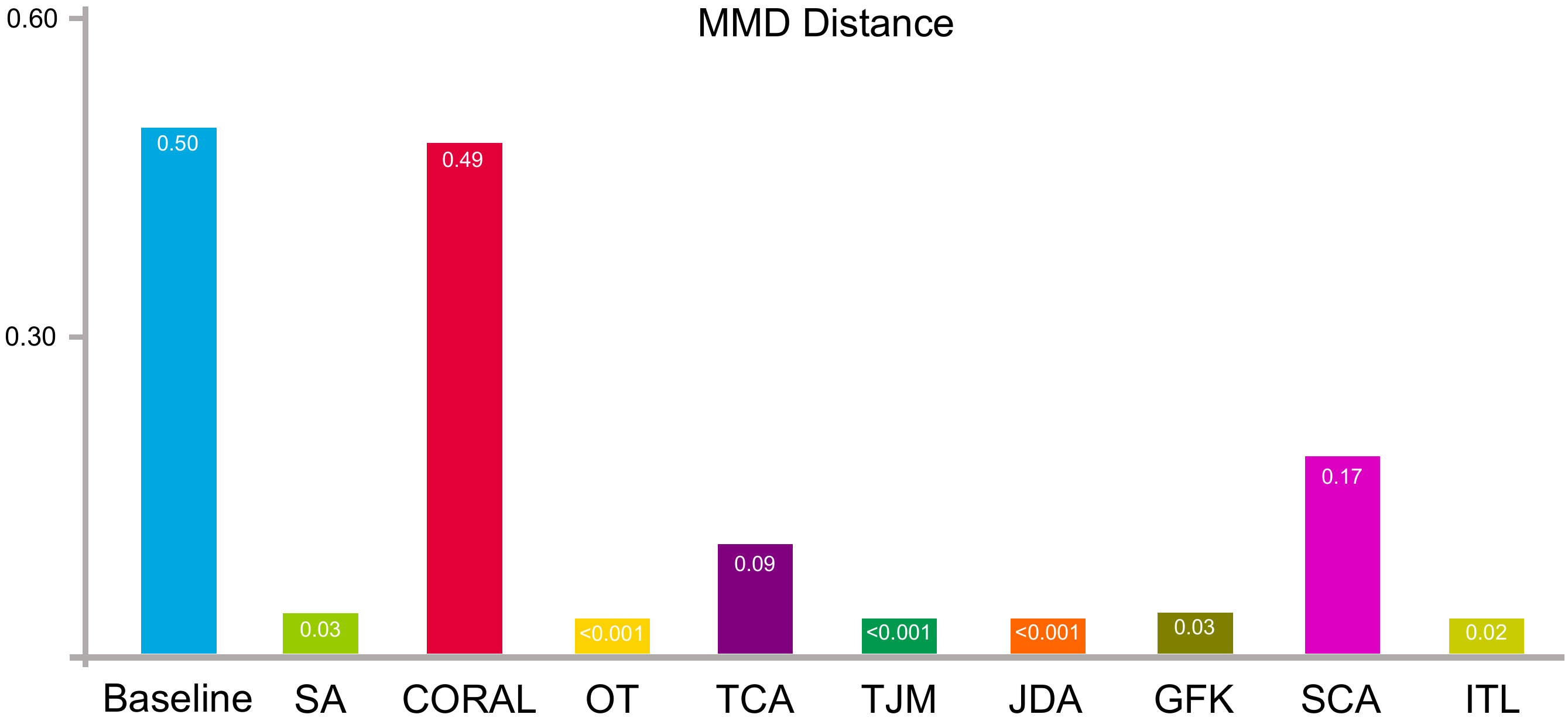}
 \caption{Synthetic data distribution differences in terms of maximum mean discrepancy before (baseline) and after domain adaptation using nine feature-level adaptation methods.}
 \label{fig_Synthetic_MMD}
\end{figure}

\begin{figure}[t]
\center
 \includegraphics[width= 1.0\linewidth]{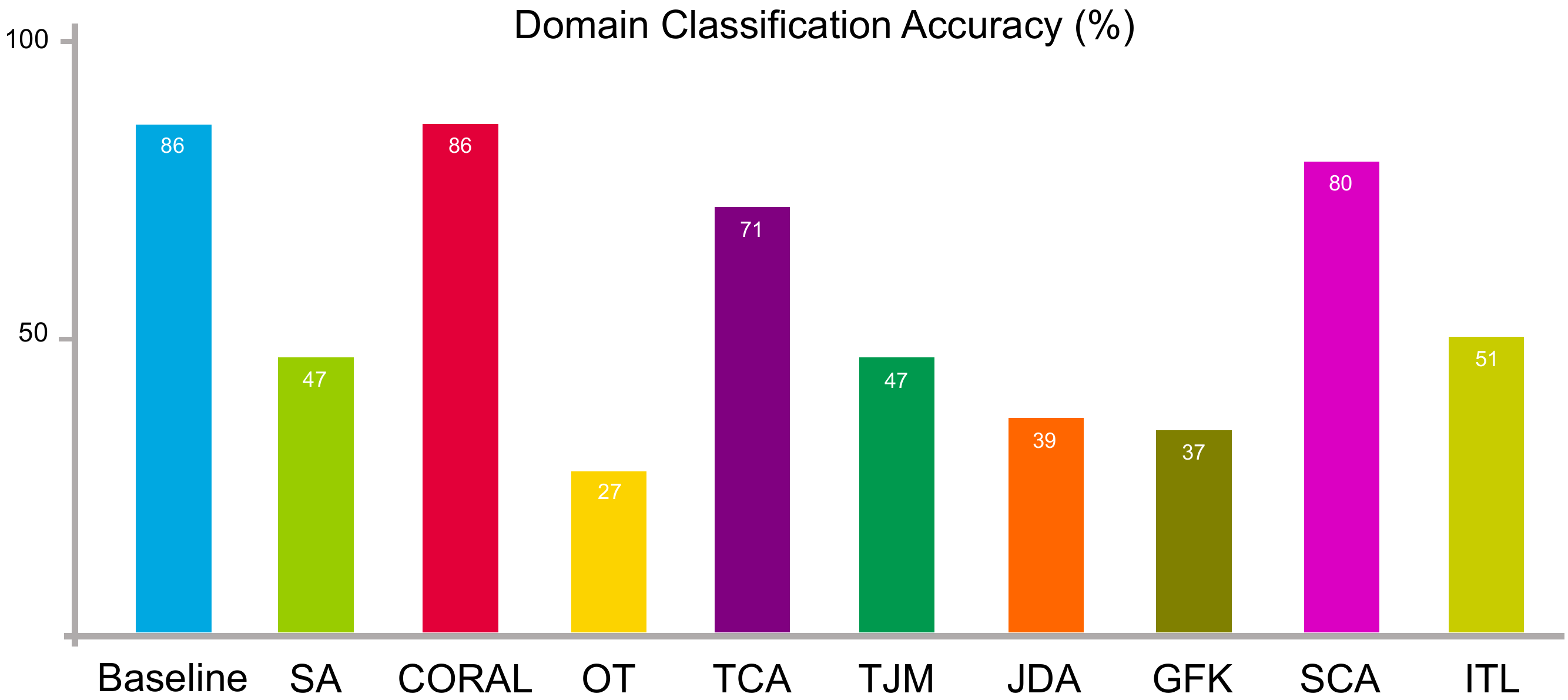}
 \caption{Synthetic data distribution differences in terms of domain-level classification accuracy on the synthetic dataset before (baseline) and after domain adaptation using nine feature-level adaptation methods.}
 \label{fig_domain_cls_synthetic}
\end{figure}
\begin{figure*}[t]
\center
 \includegraphics[width= 1.0\linewidth]{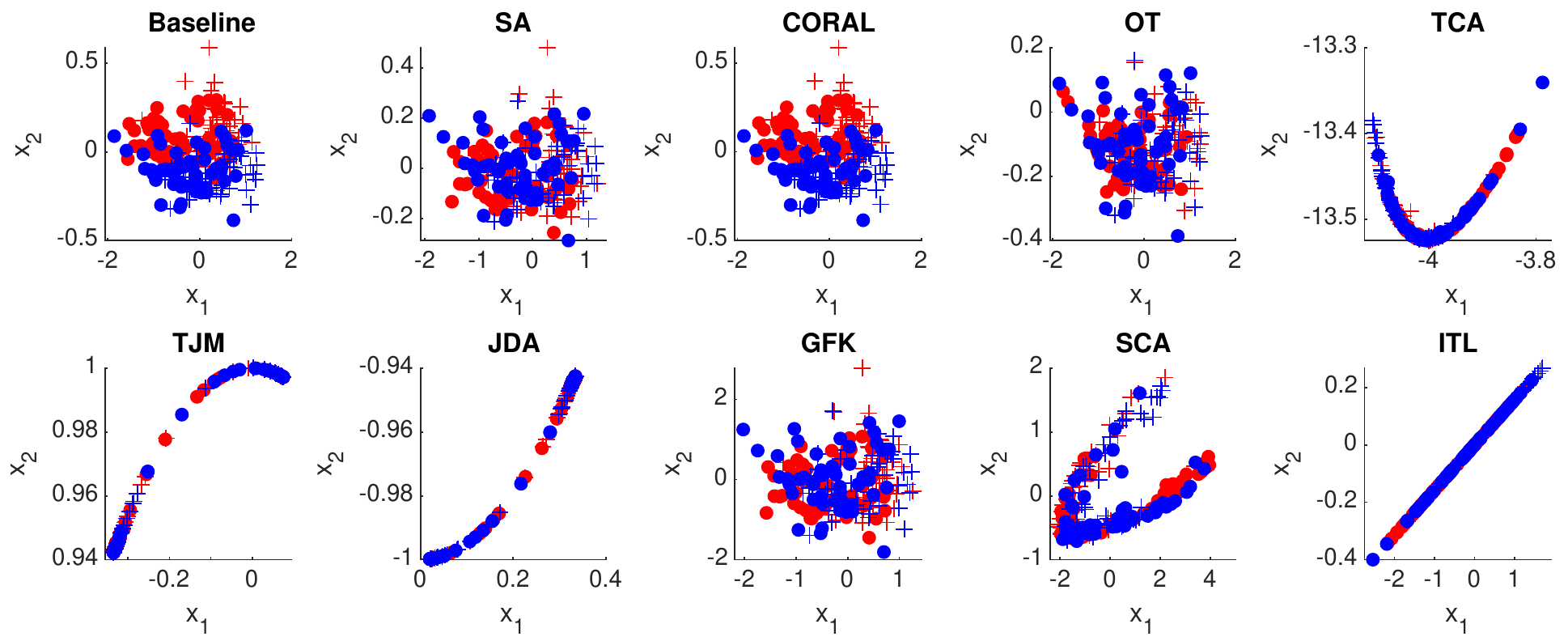}
 \caption{Distribution of the original ADNI data (baseline) and adapted data by nine feature-level domain adaptation methods in the DomainATM toolbox. ({\color{red}\textbf{+}} positive source samples; {\color{blue}\textbf{+}} positive target samples; {\color{red}\textbf{$\bullet$}} negative source samples; {\color{blue}\textbf{$\bullet$}} negative target samples)}
 \label{fig_ADNI_dis}
\end{figure*}
\begin{figure}[!t]
\center
 \includegraphics[width= 1.0\linewidth]{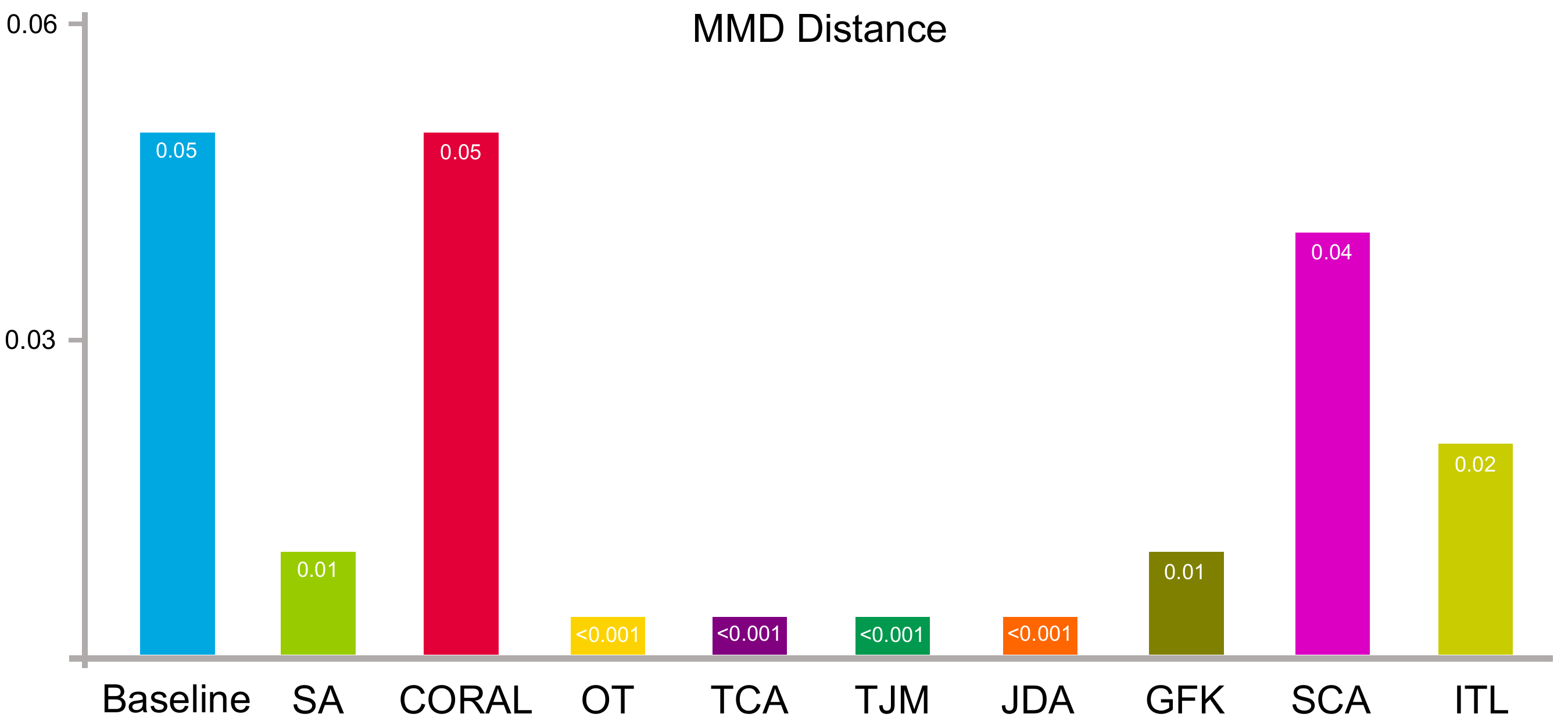}
 \caption{Data distribution differences in terms of maximum mean discrepancy on ADNI-1 and ADNI-2 before (baseline) and after domain adaptation operations.}
 \label{fig_ADNI_MMD}
\end{figure}

\begin{figure}[!t]
\center
 \includegraphics[width= 1.0\linewidth]{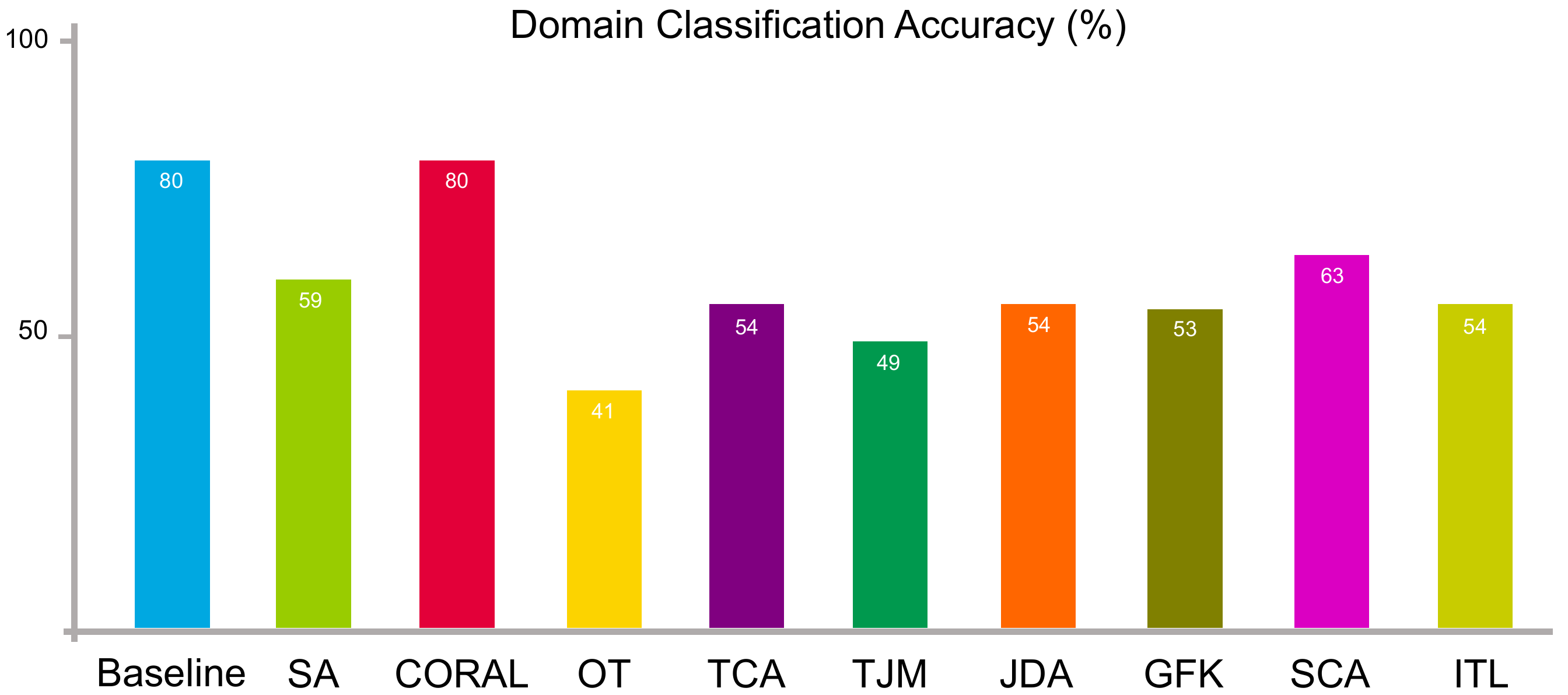}
 \caption{Data distribution differences in terms of domain-level classification accuracy on ADNI-1 and ADNI-2 before (baseline) and after domain adaptation operations.}
 \label{fig_ADNI_site_cls}
\end{figure}
\begin{figure*}[t]
\center
 \includegraphics[width= 1.0\linewidth]{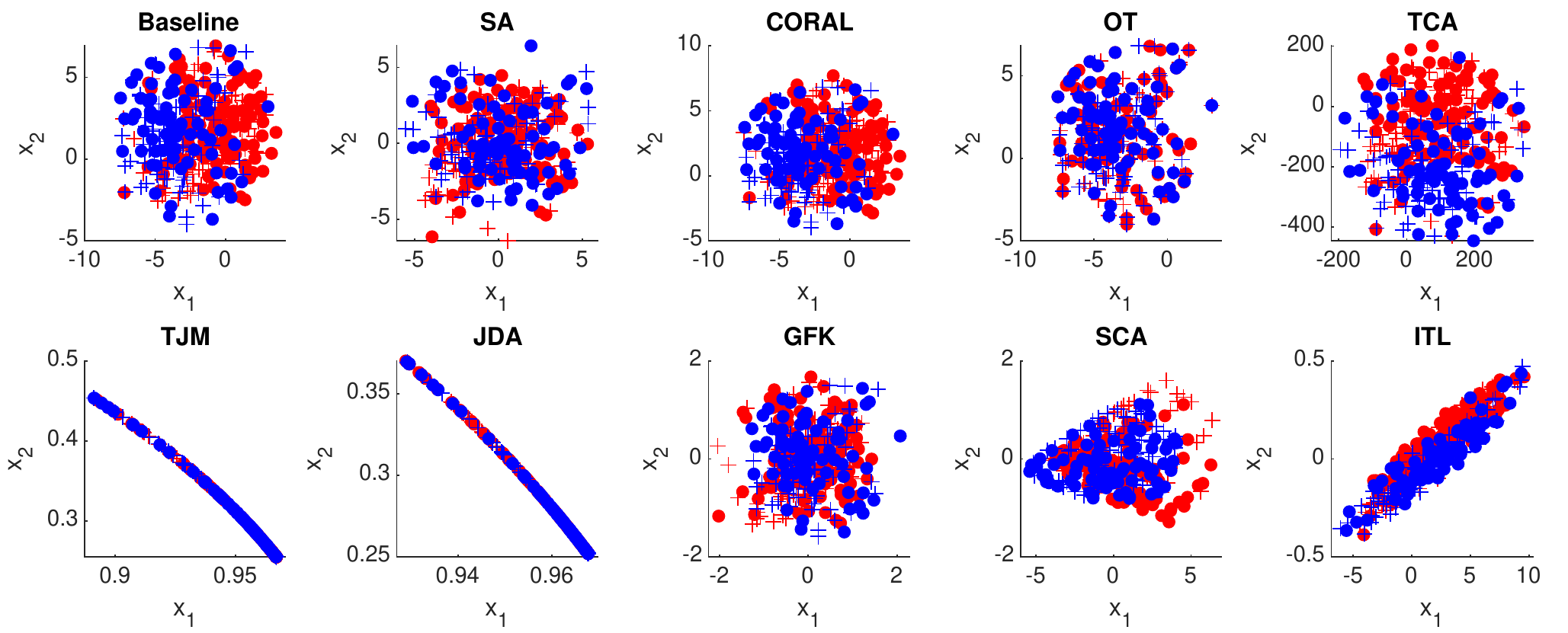}
 \caption{Distribution of the original ABIDE data (baseline) and adapted data by nine feature-level domain adaptation methods in the proposed DomainATM toolbox. ({\color{red}\textbf{+}} positive source samples; {\color{blue}\textbf{+}} positive target samples; {\color{red}\textbf{$\bullet$}} negative source samples; {\color{blue}\textbf{$\bullet$}} negative target samples)
 }
 \label{fig_ABIDE_dis}
\end{figure*}

\begin{figure}[!t]
\center
 \includegraphics[width= 1.0\linewidth]{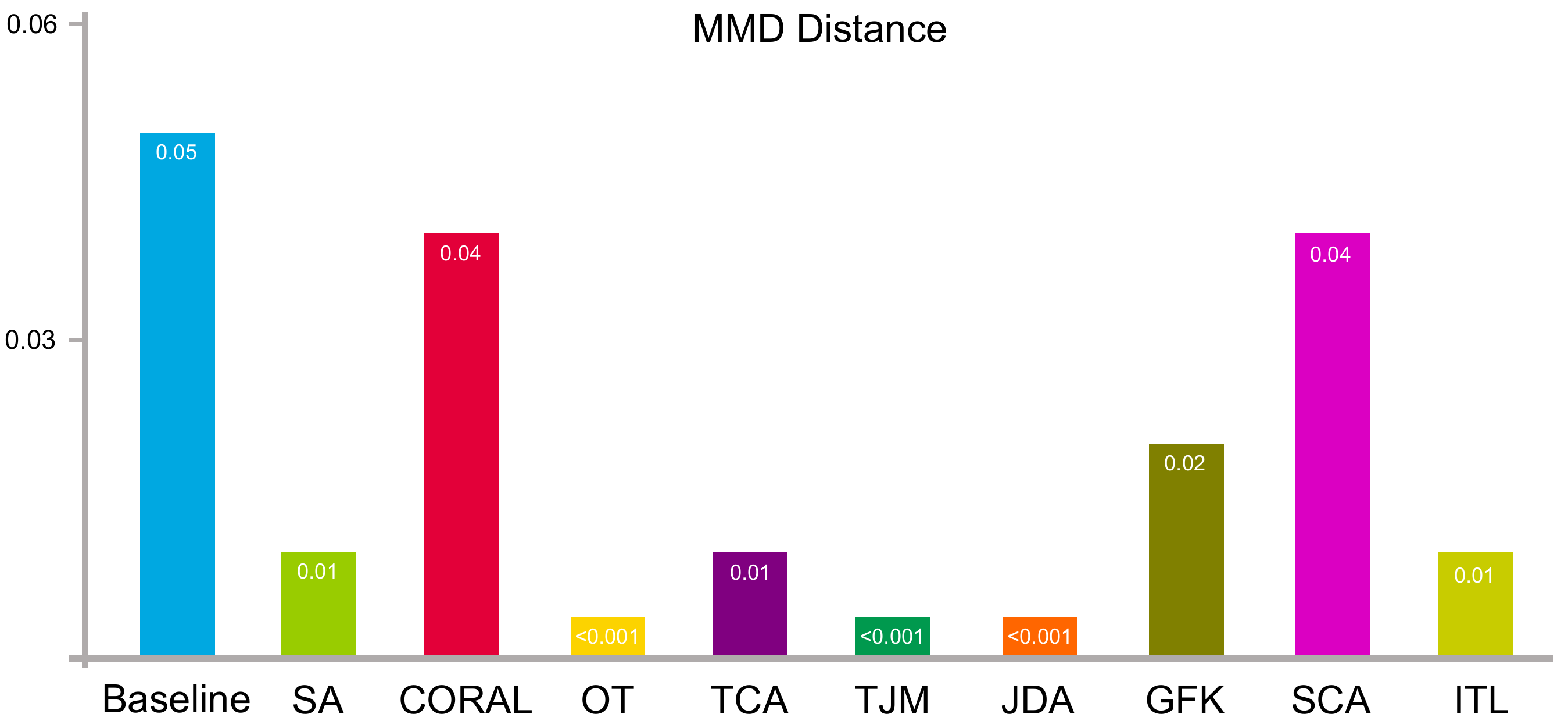}
 \caption{Data distribution differences of two sites of ABIDE in terms of maximum mean discrepancy before (baseline) and after domain adaptation using nine feature-level adaptation methods.}
 \label{fig_ABIDE_MMD}
\end{figure}

\begin{figure}[!t]
\center
 \includegraphics[width= 1.0\linewidth]{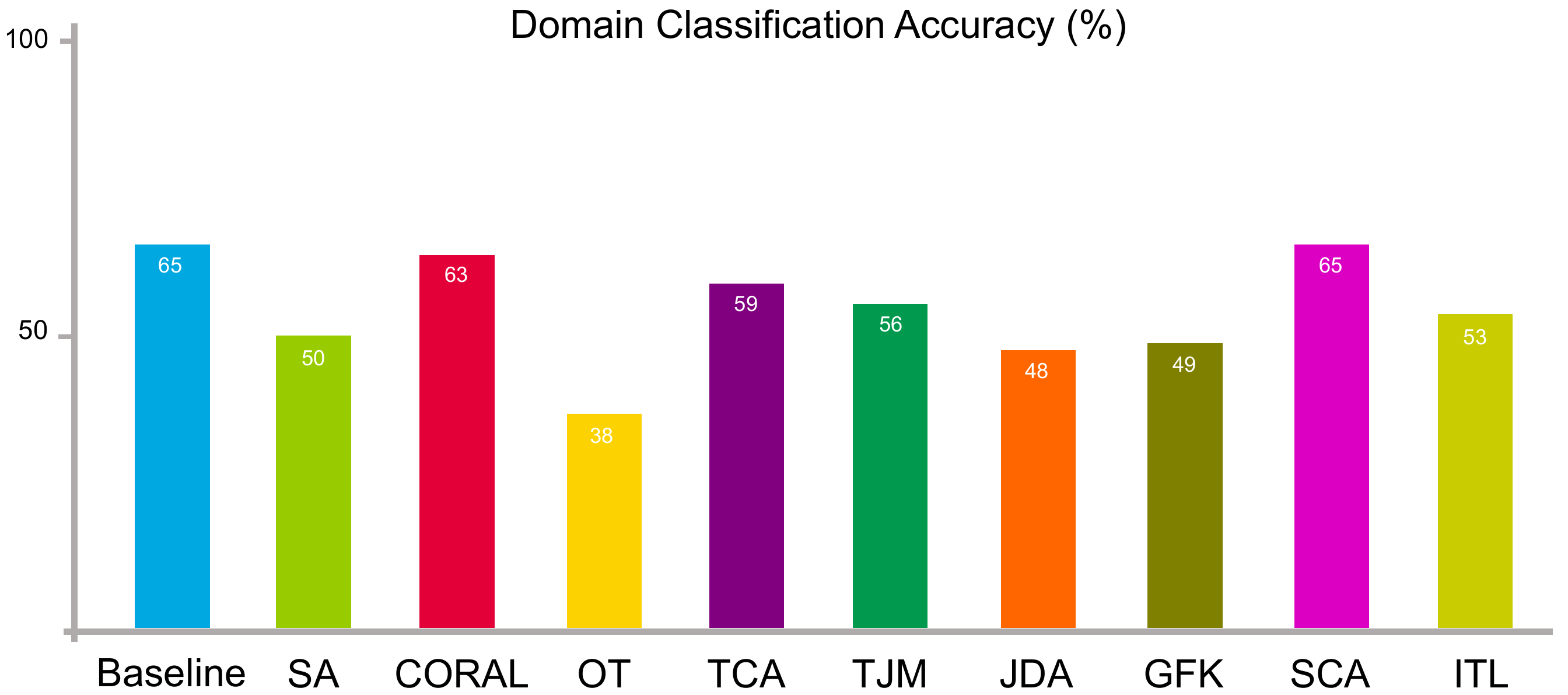}
 \caption{Data distribution differences in terms of domain-level classification accuracy on two sites of ABIDE before (baseline) and after domain adaptation using nine feature-level adaptation methods.}
 \label{fig_ABIDE_site_cls}
\end{figure}

\section{Empirical Evaluation of Feature-Level Data Adaptation Algorithms in DomainATM} 
\label{Experiment1}
\subsection{Evaluation Metric}
For feature-level adaptation methods, we adopt the metrics that evaluate the distribution changes before and after the adaptation process.
Specifically, we use the following three methods/metrics for adaptation performance evaluation.

\begin{itemize}
\item \textbf{Distribution Difference.}
We adopt maximum mean discrepancy (MMD) to measure the data distribution differences of the source and target domains before and after domain adaptation. 
As a popular metric, the maximum mean discrepancy (MMD) has been widely used in domain adaptation research~\cite{TCA,JDA,TJM, MMD-1,MMD-2,MMD-3}, defined as follows: 
\begin{equation}
\mathbf{MMD}^2_k = \left \| \mathbf{E}_p [\phi (\mathbf{x}^s)  ] -  \mathbf{E}_q [\phi (\mathbf{x}^t)  ]               \right\|^2_{\mathcal{H}_k}
\label{mmd_loss}
\end{equation}
where ${\mathcal{H}_k}$ denotes the Reproducing Kernel Hilbert Space endowed with a kernel function $k$, and
$k(\mathbf{x}^s, \mathbf{x}^t) =\left \langle \phi (\mathbf{x}^s), \phi (\mathbf{x}^t) \right \rangle$.
If the MMD distance of source and target domains gets lower after adaptation, it indicates the data distribution difference becomes smaller.

\item \textbf{Domain Classification.}
Suppose an equal number of samples are sampled from the source and target domains, respectively. These samples are assigned with {\em domain labels}, \ie, the source samples are labeled as ``1" while target samples are assigned with the label ``0". 
A {\em domain discriminator/classifier} is applied to all these samples for  distinguishing which samples come from the source domain and which ones are from the target domain. The classification result is used to assess domain shift/difference. 
A high domain classification accuracy indicates that the source and target samples can be easily distinguished, which means the domain shift is large. 
On the contrary, if the domain classification accuracy drops down after the adaptation processing, it indicates the domain adaptation algorithm works because it makes the two domains get closer and become more difficult to distinguish.

\if false
\item \textbf{Category classification analysis.}
Train a classifier on the source data with their category labels. Then apply the trained classifier to the target domain to facilitate category classification. 
To assess the domain adaptation performance, we calculate the classification metrics before and after the adaptation processing. If the classification scores increase, it indicates that the adaptation algorithm works. In practice, we adopt accuracy (ACC), sensitivity (SEN), specificity (SPE) and AUC as the classification metrics.
\fi

\end{itemize}

\subsection{Experiment 1: Adaptation on Synthetic Dataset}
We first conduct experiments on synthetic datasets using DomainATM.
Specifically, we generate two domains by Gaussian distributions. 
Each domain has two classes, with 30 positive samples and 30 negative ones, respectively.
For the source domain $\mathcal{S}$, the means of positive and negative samples are [0, 0] and [0, 1], while their covariance matrices are [0.2, 0; 0, 0.2] and [0.1, 0; 0, 0.1].
For the target domain $\mathcal{T}$, the means of positive and negative samples are [1, -0.5] and [1, 0.2], while their covariance matrices are [0.2, 0; 0, 0.2] and [0.1, 0; 0, 0.1].

\subsubsection{Data Distribution Visualization}
The distributions of the original data and the adapted data by different methods are visualized in Fig~\ref{fig_synthetic}.
From the visualization result, different domain adaptation methods can reduce the distributions of source and target samples to certain extents.
For example, the optimal transport adaptation (OT) can project the source data into the target domain, and make the source distribution quite similar to the target domain.
\subsubsection{Distribution Difference}
The data distribution differences (in terms of maximum mean discrepancy) of the source and target domains after domain adaptation are shown in Fig.~\ref{fig_Synthetic_MMD}.
The result of the Baseline method shows the original distribution of the source and target domain without any adaptation processing.
From Fig.~\ref{fig_Synthetic_MMD}, we can observe that domain adaptation can reduce the distribution differences between the original source and target domains.
\subsubsection{Domain-Level Classification}
We conduct domain-level classification on the source and target data. A domain classifier (we use a k-nearest neighbors classifier) is trained with source data (with label ``1'') and target data (with label ``0''). 
Source and target data are combined together and shuffled. 
In the experiments, we use 60\% of the entire data samples for training the domain classifier while 40\% are for test. 
The result of domain classification accuracy is shown in Fig.~\ref{fig_domain_cls_synthetic}.
From the result, it can be seen that the domain classification accuracy drops after domain adaptation. This implies that the source and target data become more difficult to be distinguished, \ie, domain adaptation makes their distributions become more similar than in the original data space. 


\subsection{Experiment 2: Adaptation for Alzheimer's Disease Analysis on ADNI}

We conduct experiments on the Alzheimer's Disease Neuroimaging Initiative (ADNI) dataset~\cite{ADNI}. The dataset consists of structural brain MRI data for Alzheimer's disease analysis. We use two subsets of ADNI, \ie, ADNI-1 (100 subjects) and ADNI-2 (100 subjects) as the source and target domains, respectively, to test the domain adaptation algorithms using DomainATM. 
All the MRIs have been processed through a standard pipeline, including skull stripping, intensity correction, registration and re-sampling. Regions-of-interest (ROIs) features which are defined on 90 regions in the Anatomical Automatic Labeling (AAL) atlas~\cite{AAL} are used to represent each subject. 

\subsubsection{Distribution Visualization}
The distributions of the original ADNI-1 and ADNI-2 data (in feature space) and the adapted data by different methods are visualized in Fig~\ref{fig_ADNI_dis}.
From the visualization results, the original source and target data have a relatively clear boundary. After domain adaptation processing, the domain boundaries become blurred, and the distribution of source and target domains gets closer to each other.

\subsubsection{Distribution Distance}
The data distribution differences (in terms of maximum mean discrepancy) of the source domain, \ie, ADNI-1, and target domains, \ie, ADNI-2, after domain adaptation are shown in Fig.~\ref{fig_ADNI_MMD}.
The baseline illustrates the original distribution of the source and target domain without any adaptation processing.
From the result, it can be observed that domain adaptation is able to reduce the distribution differences between the original source and target domains.

\subsubsection{Domain-Level Classification}
We facilitate domain-level classification on the source data, \ie, ADNI-1, and target data, \ie, ADNI-2. A domain classifier (k-nearest neighbors classifier) is trained with source data (with label ``1'') and target data (with label ``0''). 
Source and target data are combined together and shuffled. 60\% of the entire data are adopted for training while 40\% for testing. 
The result of domain-level classification is illustrated in Fig.~\ref{fig_ADNI_site_cls}.
From the result, we can see that the domain classification accuracy drops after domain adaptation. This indicates that the adapted source and target data get
more difficult to be classified, \ie, domain adaptation is effective in reducing their distribution differences.

\begin{table*}[!tbp]
\linespread{1.4}
\caption{Results of three image-level domain adaptation methods on T1-weighted MRIs of five travelling phantom subjects acquired by three different scanners from the ABCD dataset.}
\begin{center}
\setlength{\tabcolsep}{5mm}{
\begin{tabular} {l | l | c |c |c } 
\toprule[1.2pt]
Source Domain$\rightarrow$Target Domain (Subject ID) &Method     &CC     &PSNR       &MSE \\

\toprule

\multirow{3}*{GE$\rightarrow$Siemens (Phantom-2, Phantom-3)}     
&Baseline   &0.4889$\pm$0.0081   &21.4143$\pm$2.8718    &0.0080$\pm$0.0049\\ 
&HM          &0.5642$\pm$0.0395  &22.3131$\pm$2.5975    &0.0064$\pm$0.0037\\
&SSIMH   &0.5935$\pm$0.0221   &22.7624$\pm$2.5310    &0.0057$\pm$0.0032\\ 

\midrule

\multirow{3}*{Philips$\rightarrow$Siemens (Phantom-4, Phantom-5)}    
&Baseline   &0.5408$\pm$0.0194   &18.7578$\pm$0.8847  &0.0135$\pm$0.0028\\
&HM         &0.5495$\pm$0.0388     &18.7477$\pm$1.1303  &0.0135$\pm$0.0035\\
&SSIMH    &0.6098$\pm$0.0269    &20.1269$\pm$1.8421  &0.0101$\pm$0.0042\\ 

\midrule

\multirow{3}*{GE$\rightarrow$Philips (Phantom-1)}    
&Baseline     &0.4682     &21.3915      &0.0073 \\
&HM           &0.5108     &21.2482      &0.0075 \\
&SSIMH           &0.5570     &22.6421      &0.0054 \\ 
\bottomrule[1.2pt]
\end{tabular}
}
\end{center}

\label{tab_T1}
\end{table*}

\subsection{Experiment 3: Domain Adaptation for Autism Analysis on ABIDE}

We conduct experiments on the Autism Brain Imaging Data Exchange (ABIDE) dataset~\cite{ABIDE}. This database consists of rest-state functional MRI data for Autism analysis. 
We use two sites from the ABIDE project, \ie, Leuven (57 subjects) and USM (60 subjects) as the source and target domains, respectively, to test the domain adaptation algorithms using the DomainATM. 
All the fMRIs go through a standard pipeline, including slice-timing and motion correction, nuisance signal regression, temporal filtering, and registration.
Mean time series of 116 regions-of-interest (ROIs) defined by the Anatomical Automatic Labeling (AAL) atlas~\cite{AAL} are extracted. 
Then a $116\times116$ symmetrical resting-state functional connectivity (FC) matrix is generated for each subject, with each element representing the Pearson correlation coefficient between a pair of ROI signals.
We use some graph feature (\ie, betweenness centrality) based on the FC matrix to represent each subject/sample.
\subsubsection{Distribution Visualization}
The original distributions of two sites in ABIDE (in feature space) and the adapted data by different methods are visualized in Fig~\ref{fig_ABIDE_dis}.
From the visualization result, it can be observed that the boundary of original source and target data is relatively clear. After the domain adaptation processing, the domain boundaries become blurred, and the distributions of source and target domain get similar to each other.

\subsubsection{Distribution Distance}
The data distribution differences (in terms of maximum mean discrepancy) of the source domain, \ie, NYU, and target domain, \ie, UM, after domain adaptation are shown in Fig.~\ref{fig_ABIDE_MMD}.
The baseline is the original distribution of the source and target domain without any adaptation processing.
The result shows that the distribution differences become smaller after adaptation processing by different algorithms.

\subsubsection{Domain-Level Classification}
We facilitate domain-level classification on the source data, \ie, NYU, and target data, \ie, UM. A domain classifier (k-nearest neighbors classifier) is trained with source data (with label ``1'') and target data (with label ``0''). 
Source and target data are combined together and shuffled. 60\% of the entire data are adopted for training while 40\% for test. 
The result of domain-level classification accuracy is illustrated in Fig.~\ref{fig_ABIDE_site_cls}.
From Fig.~\ref{fig_ABIDE_site_cls}, the domain classification accuracy gets worse after domain adaptation processing. This indicates that the adapted source and target data become more difficult to be discriminated, \ie, using domain adaptation has successfully reduced their distribution differences. 


\begin{figure*}[!tbp]
\center
 \includegraphics[width= 1.0\linewidth]{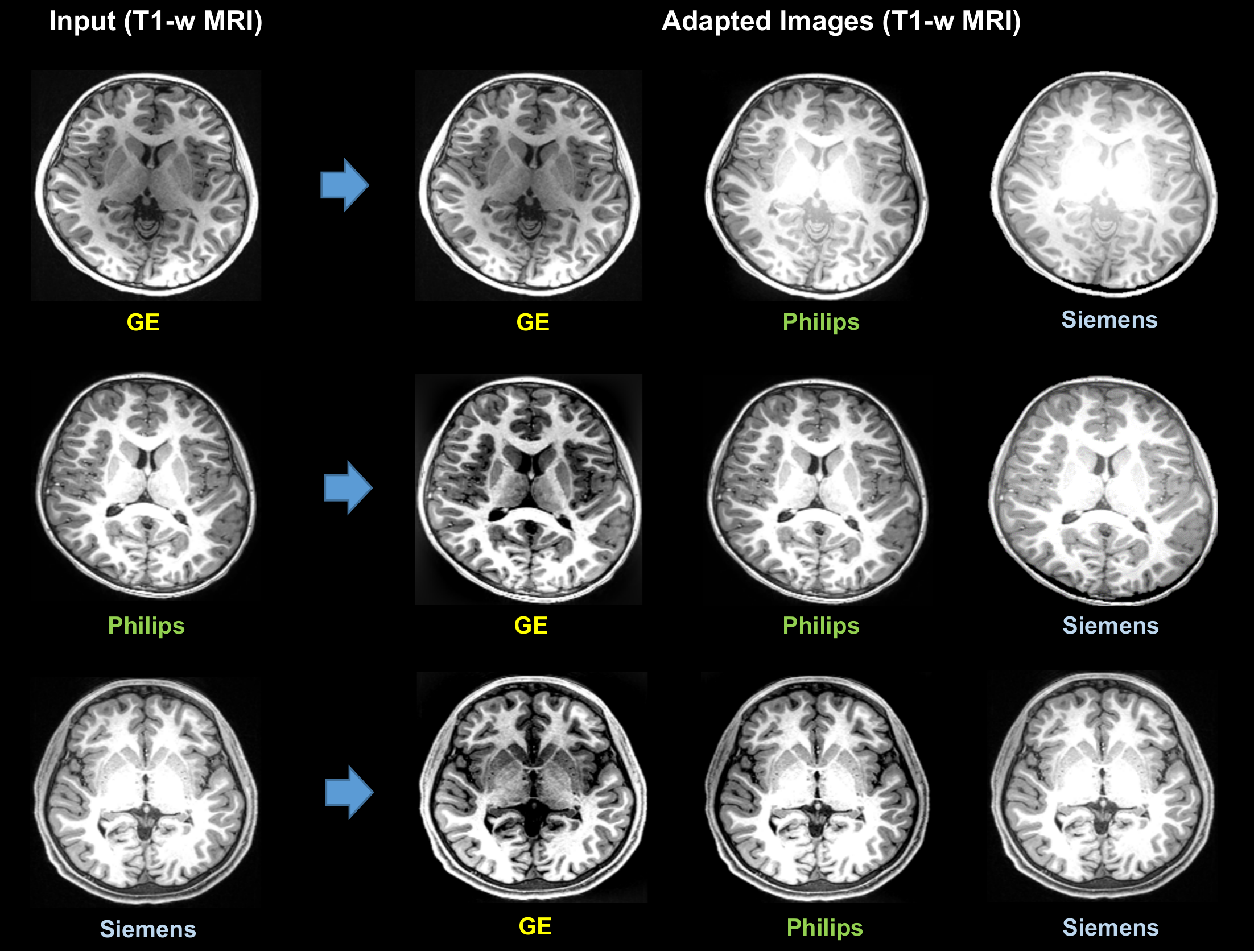}
 \caption{Image-level domain adaptation via the Spectrum Swapping-based Image-level Harmonization (SSIMH) method~\cite{SSIMH} for T1-weighted (T1-w) MRIs acquired by different scanners. Domain shift caused by the use of different scanners can be partly reduced by image-level adaptation via SSIMH.}
 \label{fig_GPS}
\end{figure*}
\section{Empirical Evaluation of Image-Level Data Adaptation Algorithms in DomainATM}
\label{Experiment2}

\subsection{Evaluation Metrics}
For image-level adaptation methods, we adopt the metrics that evaluate the image similarity/dissimilarity before and after adaptation.
Specifically, we adopt the following three metrics for image-level adaptation performance evaluation.

\begin{itemize}
\item \textbf{Correlation Coefficient (CC).}
Denote the source and target images as $\mathcal{I}_s$ and $\mathcal{I}_t$. 
After the adaptation, we get $\mathcal{I}_s'$. 
For adaptation performance assessment, if the correlation coefficient of $\mathcal{I}_s'$ and $\mathcal{I}_t$ is higher than $\mathcal{I}_s$ and $\mathcal{I}_t$, it indicates the corresponding adaptation algorithm works.

\item \textbf{Peak Signal-to-Noise Ratio (PSNR).}
If the peak signal-to-noise ratio of $\mathcal{I}_s'$ and $\mathcal{I}_t$ is higher than $\mathcal{I}_s$ and $\mathcal{I}_t$, it indicates the adaptation algorithm works.

\item \textbf{Mean-Squared Error (MSE).}
If the mean-squared error of $\mathcal{I}_s'$ and $\mathcal{I}_t$ is smaller than $\mathcal{I}_s$ and $\mathcal{I}_t$, it indicates the adaptation algorithms is effective.

\end{itemize}
\subsection{Materials and Settings}
Phantom data of five traveling subjects with T1-weighted (T1-w) structural MRIs from the ABCD dataset~\cite{ABCD} are used for performance evaluation. 
Phantom-1 is scanned by GE and Philips scanners, respectively.
Phantom-2 and Phantom-3 are acquired by Siemens and GE scanners, respectively.
Phantom-4 and Phantom-5 are scanned by Philips and Siemens scanners, respectively.
These phantoms are used to test the performance of image-level domain adaptation methods in handling domain shift caused by different scanners.

All these 3D MRIs are raw data in the {\em NIfTI} file format. We do not perform any pre-processing such as skull-stripping, registration or segmentation before image-level adaptation. 
During adaptation, the intensity of each image is normalized to the range of [0, 1]. 
For these volumetric images which contain multiple slices, the adaptation is facilitated on each slice, then the performance is calculated as an average metric value for all the slices within an image (volume).

\subsection{Result}
We conduct image-level domain adaptation on these five phantom structural MRI data, and the adaptation results in terms of the three metrics are shown in Table~\ref{tab_T1}.
From the result, it can be observed that image-level domain adaptation methods can generally achieve higher scores of correlation coefficient (CC) and peak signal-to-noise ratio (PSNR) and smaller mean square error (MSE). In some cases (\eg, GE $\rightarrow$ Philips), the Histogram Matching (HM) does not perform very well in terms of PSNR and MSE.
Overall, the result indicates that the image-level adaptation methods are useful in reducing the distribution shift between images caused by different imaging scanners.

\subsection{Visual Inspection}
To further investigate the effectiveness of image-level domain adaptation, we do visual inspections of the MRIs that are adapted to different scanner styles.
We divide the phantom MRIs into three groups in terms of the canners.
Then we adapt MRIs acquired by one scanner to the styles of MRIs scanned by other scanners. 
We use the SSIMH method~\cite{SSIMH} in DomainATM to perform image-level adaptation.
Fig.~\ref{fig_GPS} shows the results of three different MRIs and their corresponding adapted images to different scanner styles. 
From the result, we have the following two observations. 1) Different scanners, \ie, Siemens, Philips, GE, have significant impact on the MRIs, which can cause the domain shift. 
2) The image-level domain adaptation method is effective in harmonizing the source image to the target image (reference image), and reduce the domain shift caused by different scanners.

\section{Conclusion and Future Work} \label{Conclusion}
Domain adaptation has become an important topic in the field of medical data analysis.
In this paper, we develop a Domain Adaptation Toolbox for Medical data analysis (DomainATM), aiming to help researchers facilitate fast domain adaptation for medical data acquired from different sites/scanners. 
The DomainATM is easy to use, efficient to run, and most importantly, it is able to do both feature-level and image-level adaptation.
In addition, users can add their own domain adaptation algorithms into the toolbox, making it flexible and extensible.
Experiments on both synthetic and real-world medical datasets have been conducted to show the usage and effectiveness of DomainATM.
We hope the toolbox can provide more convenience and benefit for researchers to do domain adaptation research in medical data analysis.

There are several potential future works to further enrich and extend the DomainATM. 
We will incorporate more domain adaptation algorithms and domain-oriented data harmonization methods, such as GAN~\cite{GAN-1,GAN-2} and ComBat~\cite{ComBat,NeuroCombat,ComBat0}.
Besides, we will further improve the graphic user interface to enable users to set and tune the hyper-parameters of each domain adaptation method in a more convenient manner.

\if false
\section*{Acknowledgment}
This work was supported in part by NIH grants (No.~AG073297 and AG041721).
\fi


\footnotesize
\bibliographystyle{IEEEtran}
\bibliography{mybib}

\end{document}